\documentclass{article}

 \usepackage[main, final]{neurips_2025}

\usepackage[utf8]{inputenc} %
\usepackage[T1]{fontenc}    %
\usepackage{hyperref}
 \hypersetup{
     colorlinks=true,
     linkcolor=blue,
     filecolor=blue,
     citecolor = black,      
     urlcolor=blue,
     }
\usepackage{url}            %
\usepackage{booktabs}       %
\usepackage{amsfonts}       %
\usepackage{nicefrac}       %
\usepackage{microtype}      %
\usepackage{xcolor}         %
\usepackage{comment}
\usepackage{amsmath}
\newcommand{\newc}{\newcommand}
\newc{\E}{\mathbb{E}}
\newc{\V}{\mbox{V}}
\newc{\N}{\mbox{N}}
\newc{\Bern}{\mbox{Bern}}
\newc{\Po}{\mbox{Po}}
\newc{\IG}{\mbox{IG}}
\newc{\Gam}{\mbox{Gam}}
\newc{\bdp}{\mathbf{p}}
\newc{\bdt}{\mathbf{t}}
\newc{\Normal}{\mathcal{N}}
\newc{\Expectation}{\mathbb{E}}
\newc{\odds}{\mbox{odds}}
\newc{\OR}{\mbox{OR}}
\newc{\stderr}{\mbox{s.e.}}
\newc{\logit}{\mbox{logit}}
\newc{\sign}{\mbox{sign}}
\newc{\SD}{\mbox{SD}}
\newc{\deltahat}{\hat{\delta}}
\newc{\shat}{\hat{s}}
\newc{\bdq}{\mathbf{q}}
\newc{\bdmu}{\mbox{\boldmath $\mu$}}
\newc{\bdSigma}{\mbox{\boldmath $\Sigma$}}
\newc{\bdLambda}{\mbox{\boldmath $\Lambda$}}
\newc{\bdmuhat}{\mbox{\boldmath $\hat{\mu}$}}
\newc{\bdeta}{\mbox{\boldmath $\eta$}}
\newc{\bdtheta}{\mbox{\boldmath $\theta$}}
\newc{\bdbeta}{\mbox{\boldmath $\beta$}}
\newc{\bdgamma}{\mbox{\boldmath $\gamma$}}
\newc{\bdbetahat}{\mbox{\boldmath $\hat{\beta}$}}
\newc{\bdgammahat}{\mbox{\boldmath $\hat{\gamma}$}}
\newc{\bdthetahat}{\mbox{\boldmath $\hat{\theta}$}}
\newc{\bdvareps}{\mbox{\boldmath $\varepsilon$}}
\newc{\bdzero}{\mbox{\boldmath $0$}}
\newc{\bdone}{\mbox{\boldmath $1$}}
\newc{\bdnu}{\mbox{\boldmath $\nu$}}
\newc{\bdell}{\mbox{\boldmath $\ell$}}
\newc{\bdxi}{\mbox{\boldmath $\xi$}}
\newc{\bdomega}{\mbox{\boldmath $\omega$}}
\newc{\bdepsilon}{\mbox{\boldmath $\varepsilon$}}
\newc{\bdI}{\mathbf{I}}
\newc{\bdP}{\mbox{\boldmath $P$}}
\newc{\bdX}{\mbox{\boldmath $X$}}
\newc{\bdA}{\mbox{\boldmath $A$}}
\newc{\bdB}{\mbox{\boldmath $B$}}
\newc{\bdC}{\mbox{\boldmath $C$}}
\newc{\bdD}{\mbox{\boldmath $D$}}
\newc{\bdG}{\mbox{\boldmath $G$}}
\newc{\bdJ}{\mbox{\boldmath $J$}}
\newc{\bdK}{\mbox{\boldmath $K$}}
\newc{\bda}{\mbox{\boldmath $a$}}
\newc{\bdb}{\mbox{\boldmath $b$}}
\newc{\bdc}{\mbox{\boldmath $c$}}
\newc{\bde}{\mbox{\boldmath $e$}}
\newc{\bdu}{\mbox{\boldmath $u$}}
\newc{\bdv}{\mbox{\boldmath $v$}}
\newc{\bdx}{\mbox{\boldmath $x$}}
\newc{\bdy}{\mbox{\boldmath $y$}}
\newc{\bdz}{\mbox{\boldmath $z$}}
\newc{\bdr}{\mbox{\boldmath $r$}}
\newc{\bdQ}{\mbox{\boldmath $Q$}}
\newc{\bdR}{\mbox{\boldmath $R$}}
\newc{\bdY}{\mbox{\boldmath $Y$}}
\newc{\bdT}{\mbox{\boldmath $T$}}
\newc{\bdW}{\mbox{\boldmath $W$}}
\newc{\bdH}{\mbox{\boldmath $H$}}
\newc{\bdL}{\mbox{\boldmath $L$}}
\newc{\bdU}{\mbox{\boldmath $U$}}
\newc{\bdV}{\mbox{\boldmath $V$}}
\newc{\Multinom}{\mbox{Multinom}}
\newc{\tbar}{\bar{t}}
\newc{\Var}{\mbox{Var}}
\newc{\var}{\mbox{var}}
\newc{\diag}{\mbox{diag}}
\newc{\tr}{\mbox{tr}}
\newc{\phat}{\hat{p}}
\newc{\Xbar}{\bar{X}}
\newc{\xbar}{\bar{x}}
\newc{\Ybar}{\bar{Y}}
\newc{\ybar}{\bar{y}}
\newc{\dbar}{\bar{d}}
\newc{\yhat}{\hat{y}}
\newc{\bdyhat}{\mbox{\boldmath $\hat{y}$}}
\newc{\ytil}{\tilde{y}}
\newc{\ytilde}{\tilde{y}}
\newc{\ftil}{\tilde{f}}
\newc{\Ho}{\mbox{\bf H}_o}
\newc{\Ha}{\mbox{\bf H}_a}
\newc{\phatYX}{\phat_Y - \phat_X}
\newc{\SSG}{\mbox{SSG}}
\newc{\SSB}{\mbox{SSB}}
\newc{\SSE}{\mbox{SSE}}
\newc{\SST}{\mbox{SST}}
\newc{\SSR}{\mbox{SSR}}
\newc{\SSAB}{\mbox{SSAB}}
\newc{\MSG}{\mbox{MSG}}
\newc{\MSB}{\mbox{MSB}}
\newc{\MSE}{\mbox{MSE}}
\newc{\MST}{\mbox{MST}}
\newc{\MSAB}{\mbox{MSAB}}
\newc{\dfE}{\mbox{dfE}}
\newc{\dfG}{\mbox{dfG}}
\newc{\dfB}{\mbox{dfB}}
\newc{\dfT}{\mbox{dfT}}
\newc{\dfAB}{\mbox{dfAB}}
\newc{\muhat}{\hat{\mu}}
\newc{\betahat}{\hat{\beta}}
\newc{\alphahat}{\hat{\alpha}}
\newc{\etahat}{\hat{\eta}}
\newc{\phihat}{\hat{\phi}}
\newc{\sigmahat}{\hat{\sigma}}
\newc{\cl}{\centerline}
\newc{\redtitle}[1]{ {\color{red}\und{#1}:} }
\newc{\bluetitle}[1]{ {\color{blue}\und{#1}:} }
\newc{\magentatitle}[1]{ {\color{magenta}\und{#1}:} }
\newc{\R}{\mathbb{R}}
\newc{\trans}{^\mathsf{T}}
\newc{\xtx}{\bdX\trans\bdX}
\newc{\xxtxx}{\bdX(\xtx)^{-1}\bdX\trans}
\newc{\argmin}{\operatornamewithlimits{argmin}}
\newc{\argmax}{\operatornamewithlimits{argmax}}
\newc{\setS}{\mathcal{S}}
\newc{\toP}{\overset{p}{\to}}
\newc{\toD}{\overset{d}{\to}}
\newc{\eqD}{\overset{d}{=}}
\newc{\toAS}{\overset{a.s.}{\to}}
\newc{\simIID}{\overset{\text{i.i.d.}}{\sim}}
\newc{\simIND}{\overset{\text{ind.}}{\sim}}
\newc{\thetahat}{\hat{\theta}}
\newc{\bds}{\mathbf{s}}
\newc{\thetatilde}{\tilde{\theta}}
\newc{\Phat}{\hat{P}}
\newc{\Qhat}{\hat{Q}}
\newc{\tny}{\small}
\newc{\Col}{\textnormal{Col}}
\newc{\mutilde}{\tilde{\mu}}
\newc{\vecst}{\textnormal{vec}}
\newc{\indep}{\perp \!\!\! \perp}
\newc{\Cov}{\textnormal{Cov}}
\newc{\gtilde}{\tilde{g}}
\newc{\rank}{\textnormal{rank}}
\newc{\Cor}{\textnormal{Cor}}
\newc{\Lcol}{\mathcal{L}_{col}}
\newc{\Lrow}{\mathcal{L}_{row}}
\newc{\Ell}{\mathcal{L}}
\newc{\Norm}{\mathcal{N}}
\newc{\xtilde}{\tilde{x}}
\newc{\Xtilde}{\tilde{X}}
\newc{\Atilde}{\tilde{A}}
\newc{\Xhat}{\hat{X}}
\newc{\Vhat}{\hat{V}}
\newc{\Vperp}{V^\perp}
\newc{\Rhat}{\hat{R}}
\newc{\Shat}{\hat{S}}
\newc{\uhat}{\hat{u}}
\newc{\hhat}{\hat{h}}
\newc{\Yhat}{\hat{Y}}
\newc{\bpage}{\eject}
\newc{\bdg}{\mathbf{g}}
\newc{\bdZ}{\mathbf{Z}}
\newc{\Rn}{\mathbb{R}^n}
\newc{\Sigmahat}{\hat{\Sigma}}
\newc{\Ysubj}{Y_{(j)}}
\newc{\Rsubj}{R_{(j)}}
\newc{\A}{\mathcal{A}}
\newc{\B}{\mathcal{B}}
\newc{\Gammasubj}{\Gamma_{(j)}}
\newc{\Btilde}{\tilde{\mathcal{B}}}
\newc{\Xitilde}{\tilde{\Xi}}
\newc{\xitilde}{\tilde{\xi}}
\newc{\Gammatilde}{\tilde{\Gamma}}
\newc{\gammatilde}{\tilde{\gamma}}
\newc{\Lambdatilde}{\tilde{\Lambda}}
\newc{\Ls}{\mathcal{L}}
\newc{\Lambdasubj}{\Lambda_{(j)}}
\newc{\Msubj}{M_{(j)}}
\newc{\Lsubj}{L_{(j)}}
\newc{\Ytilde}{\tilde{Y}}
\newc{\Ytildehat}{\hat{\tilde{Y}}}
\newc{\xtildebar}{\tilde{\xbar}}
\newc{\F}{\mathcal{F}}
\newc{\M}{\mathcal{M}}
\newc{\G}{\mathcal{G}}
\newc{\Dtilde}{\tilde{D}}
\newc{\Utilde}{\tilde{U}}
\newc{\Vtilde}{\tilde{V}}
\newc{\betatilde}{\tilde{\beta}}
\newc{\fhat}{\hat{f}}
\newc{\Rtilde}{\tilde{R}}
\newc{\Qtilde}{\tilde{Q}}
\newc{\gammahat}{\hat{\gamma}}
\newc{\X}{\mathcal{X}}
\newc{\Y}{\mathcal{Y}}
\newc{\abar}{\bar{a}}
\newc{\phitilde}{\tilde{\phi}}
\newc{\ahat}{\hat{a}}
\newc{\bhat}{\hat{b}}
\newc{\bdf}{\mathbf{f}}
\newc{\bdh}{\mathbf{h}}
\newc{\Pois}{\text{Pois}}
\newc{\bdlambda}{\boldsymbol{\lambda}}
\newc{\Mult}{\text{Mult}}
\newc{\liminfc}{\underline{\lim}}
\newc{\limsupc}{\overline{\lim}}
\newc{\pbar}{\overline{p}}
\newc{\bdalpha}{\boldsymbol{\alpha}}
\newc{\psibar}{\overline{\psi}}
\newc{\bdxbar}{\overline{\mathbf{x}}}
\newc{\bdd}{\mathbf{d}}
\newc{\Wtilde}{\tilde{W}}
\newc{\bdZtilde}{\tilde{\mathbf{Z}}}
\newc{\ttilde}{\tilde{t}}
\newc{\Ztilde}{\tilde{Z}}
\newc{\pihat}{\hat{\pi}}
\newc{\betahatridge}{\betahat^{\lambda}_{\text{RR}}}
\newc{\epsilonhat}{\hat{\epsilon}}
\newc{\zhat}{\hat{z}}
\newc{\qhat}{\hat{q}}
\newc{\Bhat}{\hat{B}}

\newc{\hh}{\mathcal{H}}
\newc{\Unif}{\text{Unif}}
\newc{\psihat}{\hat{\psi}}
\newc{\Expo}{\text{Expo}}
\newc{\bdxtilde}{\tilde{\mathbf{x}}}
\newc{\toDPn}{\overset{P_n}{\rightsquigarrow}}
\newc{\toDQn}{\overset{Q_n}{\rightsquigarrow}}
\newc{\Fhat}{\hat{F}}
\usepackage{ulem}
\usepackage{multirow}
\usepackage{makecell}
\usepackage{graphicx}
\usepackage{float}
\usepackage{adjustbox}

\title{Intelligently Weighting Multiple Reference Models for Direct Preference Optimization of LLMs}

\author{%
Skyler Wu$^*$\\
Stanford University\\
  \texttt{skylerw@stanford.edu}
  \And
  Aymen Echarghaoui\thanks{Equal contribution.}\\
  Stanford University\\
  \texttt{aymen20@stanford.edu} \\
}

\begin{document}

\maketitle
\vspace{-0.75cm}

\begin{abstract}
\small Fine-tuning is integral for aligning large language models (LLMs) with human preferences. Multiple-Reference Preference Optimization (MRPO) builds on Direct Preference Optimization (DPO) by fine-tuning LLMs on preference datasets while regularizing the policy towards a mixture of reference models to leverage their collective desirable properties. However, current methods for setting the reference weights are ad-hoc and statistically unsound, leading to unreliable performance. To address this, we introduce four new weighting strategies: two \textit{offline} methods that leverage held-out validation signal; one \textit{online} method that uses a sliding-window estimator to reduce overfitting; and an \textit{online} method that treats reference weighting as a $K$-armed bandit via Thompson Sampling. Experiments using Qwen2.5-0.5B as the policy model and seven reference models from the Llama, Mistral, Qwen, Yi, and Phi families (0.5B-14B each) show that \textbf{all 4 of our strategies outperform the current MRPO weighting methods} on UltraFeedback and SafeRLHF in preference accuracy. More thought-provokingly, however, we find that \textbf{single-reference DPO, using any of 6 out of 7 references, consistently outperforms all tested multiple-reference approaches} --- calling into question the practical appeal of multiple-reference approaches. \normalsize
\end{abstract}

\vspace{-0.525cm}

\section{Introduction \& Background}
\vspace{-0.25cm}

Fine-tuning is integral for aligning LLMs with human preferences, with performance gains particularly impactful for smaller models where principled incorporation of external references can significantly boost performance. Direct Preference Optimization (DPO)~\citep{rafailov2023direct} is a scalable alternative to full Reinforcement Learning from Human Feedback pipeline \citep{ziegler2019fine} by directly fine-tuning a policy model on preference pairs while regularizing to a single reference model. However, DPO's reliance on a single reference policy can limit the diversity and quality of alignment signals. Multi-Reference Preference Optimization (MRPO)~\citep{le2025multi} addresses this limitation by extending DPO to $K$ reference models. In particular, MRPO regularizes the fine-tuned policy towards a weighted mixture of the $K$ references, with weights $\bdalpha$ that govern the influence of each reference. In theory, MRPO is especially valuable when distilling complementary strengths from larger models into smaller ones. 

\textbf{However, current approaches for choosing $\bdalpha$ are statistically unsound:} they either (i) ignore heterogeneity in reference quality and task-fit, or (ii) overfit to each training data point. \textbf{To address this gap, we introduce four new weighting strategies that are more statistically robust} ---two \emph{offline} methods that leverage held-out validation signal, and two \emph{online} methods based on a sliding-window estimator and a Thompson-sampling formulation that treats reference models as bandit arms. Across UltraFeedback and SafeRLHF, all four strategies consistently outperform the current MRPO approaches on preference accuracy. However, more thought-provokingly, we find that \textbf{single-reference DPO, choosing any of 6 out of the 7 reference models, consistently outperforms all tested multiple-reference approaches} --- calling into question the practical appeal of multiple-reference approaches in general. From our extensive experiments, we provide multiple explanations for this unexpected phenomena.\footnote{Reproducible code can be found at \href{github.com/AymenEcharghaoui/intelligently-weighted-finetuning}{github.com/AymenEcharghaoui/intelligently-weighted-finetuning}.}

\section{Literature Review}
\vspace{-0.35cm}
In this section, we begin by providing an overview of Direct Preference Optimization (DPO) and its length-normalized version (LN-DPO). Next, we discuss the  Multi-Reference Preference Optimization (MRPO) framework, which is the most relevant concept to our work. More discussion of RLHF, MRPO, MDPO and Thompson Sampling is deferred to Appendix \ref{additional-literature-review}.
\vspace{-0.25cm}

\subsection{Direct Preference Optimization (DPO)}
\vspace{-0.25cm}
As introduced in \cite{rafailov2023direct}, DPO simplifies policy optimization by eliminating the explicit RL loop. It directly leverages human preference data to derive a closed-form solution for the optimal policy. This avoids fitting a separate reward model. Let $\pi_\theta(y \mid x)$ denote the fine-tuned policy
and $\pi_{\text{ref}}(y \mid x)$ the reference model. We write $(x, y^+, y^-)\!\sim\!\mathcal{D}$ for a preference dataset where $y^+$ is preferred to $y^-$ under prompt $x$. Starting from RLHF \citep{ziegler2019fine}, which maximizes the objective in Appendix \ref{additional-literature-review} Eq. \ref{eq:2} over \(\pi_\theta\) under the constraint \(\sum_y \pi_\theta(y|x) = 1\), KKT conditions yield the optimal reward and policy:
\(r_{\phi*}(x, y) = \beta \log \frac{\pi_{\theta^*}(y \mid x)}{\pi_{\text{ref}}(y \mid x)} + \beta \log Z(x),
\) 
where \(Z(x)\) is a normalizer. Substituting $r_{\phi*}$ into the Bradley-Terry model (Appendix \ref{additional-literature-review} Eq. \ref{eq:1}), DPO optimizes the following:
\begin{equation}
    \pi_{\theta^*} = \arg\min_{\theta} \mathbb{E}_{(x, y^+, y^-) \sim \mathcal{D}} \left[ - \log \sigma \left( \beta \left( \log \frac{\pi_\theta(y^+|x)}{\pi_{\text{ref}}(y^+|x)} - \log \frac{\pi_\theta(y^-|x)}{\pi_{\text{ref}}(y^-|x)} \right) \right) \right].
    \label{eq:4}
\end{equation}
\begin{comment}
Substituting the above into the Bradley-Terry model (Appendix Eq. \ref{eq:1}), we have
\begin{equation}
    p^*(y^+ \succ y^- \mid x) = \sigma\left(\log \frac{\pi_{\theta}(y^+ \mid x)}{\pi_{\text{ref}}(y^+ \mid x)} - \log \frac{\pi_{\theta}(y^- \mid x)}{\pi_{\text{ref}}(y^- \mid x)}\right),
    \label{eq:3}
\end{equation} which provides us with the following final DPO optimization problem:
\begin{equation}
    \pi_{\theta^*} = \arg\min_{\theta} \mathbb{E}_{(x, y^+, y^-) \sim \mathcal{D}} \left[ - \log \sigma \left( \beta \left( \log \frac{\pi_\theta(y^+|x)}{\pi_{\text{ref}}(y^+|x)} - \log \frac{\pi_\theta(y^-|x)}{\pi_{\text{ref}}(y^-|x)} \right) \right) \right].
    \label{eq:4}
\end{equation}
\end{comment}

When normalizing by response length under Length-Normalized Direct Preference Optimization (LN-DPO), which \citet{ahrabian2025practical} demonstrates is superior to DPO, one replaces \(\log \frac{\pi_\theta(y^\pm|x)}{\pi_{\text{ref}}(y^\pm|x)}\) with \(\frac{1}{|y^\pm|}\log \frac{\pi_\theta(y^\pm|x)}{\pi_{\text{ref}}(y^\pm|x)}\). See Appendix \ref{additional-literature-review} for details. An extension of DPO to the multi-reference setting is Multi-Direct Preference Optimization (MDPO) \citep{le2025multi} which minimizes 
\(
-\mathbb{E}_{x, y_w, y_\ell \sim \mathcal{D}}
\left[
\sum_{k=1}^{K} \alpha_k
\log\sigma\!\left(
\beta \log\frac{\pi_\theta(y^+ \mid x)}{\pi_{\mathrm{ref}}^{k}(y^+ \mid x)}
-
\beta \log\frac{\pi_\theta(y^- \mid x)}{\pi_{\mathrm{ref}}^{k}(y^- \mid x)}
\right)
\right]
\), where $\bdalpha \in \Delta^{K-1}$.

\subsection{Multi-Reference Preference Optimization (MRPO)}
\label{subsection:MRPO}
\vspace{-0.25cm}
As introduced in \citet{le2025multi}, MRPO generalizes DPO to support multiple reference policies 
\( \{ \pi_{\text{ref}}^k \}_{k=1}^K \). They extend the RLHF objective in Appendix \ref{additional-literature-review} Eq.\ref{eq:2} into a multi-reference objective (Appendix \ref{additional-literature-review} Eq.\ref{eq:mrpo-rlhf}) by using 
\(\beta \left( \sum_{k=1}^{K} \alpha_k 
D_{\mathrm{KL}}(\pi_\theta \| \pi_{\text{ref}}^k)\right)\) as regularizer, where $\bdalpha \in \Delta^{K-1}$.
\begin{comment}
\begin{equation}
\max_{\pi_\theta} \; 
\mathbb{E}_{x \sim \rho, \, y \sim \pi_\theta(\cdot|x)} [r(x, y)]
- \beta \left( 
\sum_{k=1}^{K} \alpha_k 
D_{\mathrm{KL}}(\pi_\theta \| \pi_{\text{ref}}^k)
\right),
\label{eq:5}
\end{equation}
where \( \alpha_k \) are non-negative weighting coefficients satisfying \( \sum_{k=1}^K \alpha_k = 1 \).
 Directly solving Eq.\ref{eq:mrpo-rlhf} is difficult due to the nonlinearity of \( D_{\mathrm{KL}} \), so 
they use a lower bound which yields:
\end{comment}
\cite{le2025multi} claim that MRPO consistently outperforms MDPO in many settings. Directly solving Appendix \ref{additional-literature-review} Eq.\ref{eq:mrpo-rlhf} is difficult, so 
they use a lower bound which yields:
\begin{equation}
\pi^*(y|x) \propto 
\tilde{\pi}_{\text{ref}}(y|x)
\exp\left( \frac{1}{\beta} r(x, y) \right) \quad \text{with} \quad \tilde{\pi}_{\text{ref}}(y|x) = 
\left( \sum_{k=1}^{K} \frac{\alpha_k}{\pi_{\text{ref}}^k(y|x)} \right)^{-1}.
\label{eq:6}
\end{equation}
Then, the MRPO optimization becomes the same as in Eq. \ref{eq:4} where \(\pi_{\text{ref}}\) replaced with \(\tilde{\pi}_{\text{ref}}\).
\begin{comment}
\begin{equation}
-\mathbb{E}_{(x, y^+, y^-) \sim \mathcal{D}}
\left[
\log \sigma \Big(
\beta \log \frac{\pi_\theta(y^+|x)}{\tilde{\pi}_{\text{ref}}(y^+|x)} -
\beta \log \frac{\pi_\theta(y^-|x)}{\tilde{\pi}_{\text{ref}}(y^-|x)}
\Big)\right].
\end{equation}
\end{comment}
\cite{le2025multi} proposes \textit{per-example adaptive (weighting) coefficients} based on the ``discriminative confidence" of each reference, i.e. its ability to distinguish between the outputs $y^+$ and $y^-$ :
\begin{equation}
\alpha_k =
\frac{
\big| \log \pi_{\text{ref}}^k(y^+|x) - \log \pi_{\text{ref}}^k(y^-|x) \big|
}{
\sum_{i=1}^K \big| \log \pi_{\text{ref}}^i(y^+|x) - \log \pi_{\text{ref}}^i(y^-|x) \big|
}
\label{eq:8}
\end{equation}
We denote this weighting scheme as ``Original."

\textbf{We believe that these adaptive approaches for weighting are suboptimal and statistically unprincipled.} Their first approach, uniform weighting with $\alpha_k = 1/K$, while a reasonable baseline, ignores heterogeneity across reference models: ideally, $\pi_\theta$ should be pulled more strongly towards references that most improve task performance. Their second approach, per-example weighting with $\alpha_k$ as described in Eqn. \ref{eq:8} is statistically unsound because it sets $\alpha_k$ using the same labeled pair $(x, y^+, y^-)$ that the MRPO loss will be computed on. This label-dependent weighting biases the empirical loss downward, rendering it unfaithful to the overall language-modeling task. Statistically, such a label-dependent weighting induces a generalization gap, creates a train-test mismatch, and overfits via hyperparameter leakage. \textit{To our knowledge, there has been no further work proposing weighting mechanisms for MRPO.} %

While \citet{le2025multi}'s manuscript proposes theoretical weighting schemes for general $K$, their published codebase only supports $K=2$, with their implementation involving additional $K=2$-specific streamlining on top of the theoretical formulation. Consequently, the performance of MRPO for \(K>2\) remains 
untested: the reference \(\tilde{\pi}_{\mathrm{ref}}\) may become numerically unstable.
\begin{comment}
Thus, despite its conceptual framing as a general multi-reference algorithm, the MRPO methodology has only been validated in the simplest two-reference regime, leaving its scalability and robustness for \(K>2\) untested.
\end{comment}

\section{Methods \& Analysis}
\vspace{-0.25cm}
From now on, following \citet{ahrabian2025practical}'s findings, DPO refers to LN-DPO unless otherwise stated. All methods will be implemented using length-normalization to avoid length-bias.
\vspace{-0.25cm}

\subsection{Methods}
\vspace{-0.25cm}

\textbf{Improving on MRPO, we propose $4$ new methods for intelligently selecting weighting coefficients $\alpha_k$}. Concretely, suppose we have (a) a preference dataset $\{ (\bdx_i, y^+_i, y^-_i)\}_{i=1}^N$ divided into a training split $\mathcal{D}_{\text{train}}$ and a validation split $\mathcal{D}_{\text{val}}$; (b) a policy to be fine-tuned $\pi_\theta$; and (c) reference models $\{ \pi_{\text{ref}}^k \}_{k=1}^K$. Also, suppose our fine-tuned policy model $\pi_\theta$ is optimized in mini-batches of size $B$ (i.e., one gradient step every $B$ samples). Two of our methods are offline (i.e., fixed before training), while two of our methods are online (i.e., weights are adjusted adaptively over training steps).
\vspace{-0.25cm}

\paragraph{Offline Validation-Set-Based Discrimination Weighting (VDW)} Instead of overfitting to individual training data points, one direct modification of \cite{le2025multi}'s per-example weighting approach is to set $\alpha_k$ proportional to reference model $K$'s mean discriminative confidence on the validation set, yielding offline weights $\bdalpha_{\text{discrim}} \in \Delta^{K-1}$ determined by the following:
\begin{align}
    \alpha_{\text{discrim},k} \propto \sum_{i \in \mathcal{D}_{\text{val}}} \left| \frac{\log \pi_{\text{ref}}^k(y^+_i \mid \bdx_i)}{|y_i^+|} - \frac{\log \pi_{\text{ref}}^k(y^-_i \mid \bdx_i)}{|y_i^-|}\right|.\vspace{-0.35cm}
    \label{eq:VDW}
\end{align}
This approach preserves the ``discriminative confidence" intuition of \cite{le2025multi}, while avoiding the statistical disadvantages of double-dipping into the training set.
\vspace{-0.25cm}

\paragraph{Offline Validation-Set-Based Accuracy Weighting (VAW)} Heuristically, $\pi_\theta$ should be pulled more strongly towards reference models that perform well at the desired task. As such, we propose constructing accuracy-based weights $\bdalpha_{\text{acc}} \in \Delta^{K-1}$ proportional to reference validation accuracy:
\begin{align}
    \alpha_{\text{acc}, k} \propto \text{acc}(\pi_{\text{ref}}^k; \mathcal{D}_{\text{val}}) := \sum_{i \in \mathcal{D}_{\text{val}}} \bdone\left(\frac{\log \pi_{\text{ref}}^k(y^+_i \mid \bdx_i)}{|y_i^+|} > \frac{\log \pi_{\text{ref}}^k(y^-_i \mid \bdx_i)}{|y_i^-|}\right).
    \label{eq:val-set-acc}
\end{align}
\vspace{-0.25cm}

\paragraph{Online Sliding Window Cumulative Weighting (SWCW)} Let $\mathcal{D}_{\text{train}}^{(t)}$ be the $t$th mini-batch of training data and $\pi_\theta^{(t)}$ be the MRPO fine-tuned model after gradient step $t$. For gradient step $t$, following \cite{le2025multi}'s per-example weighting intuition regarding discriminative confidence, we propose the following online variant for the weights $\bdalpha_{\text{SWCW}}^{(t)} \in \Delta^{K-1}$ at time $t$:
\begin{align}
    \alpha_{\text{SWCW}, k}^{(t)} \propto \sum_{i \in \mathcal{D}_{\text{train}}^{(t-1)}} \left| \frac{\log \pi_{\text{ref}}^k(y^+_i \mid \bdx_i)}{|y_i^+|} - \frac{\log \pi_{\text{ref}}^k(y^-_i \mid \bdx_i)}{|y_i^-|}\right|.\vspace{-0.15cm}
    \label{eq:SWCW}
\end{align}
Intuitively, after the $t-1$th gradient step, our fine-tuned model $\pi_\theta^{(t-1)}$ should be in a region of the parameter space that is more aligned with the our $K$ reference models, at least on the training data points in mini-batch $\mathcal{D}_{\text{train}}^{(t-1)}.$ Thus, when moving to the $t$th mini-batch and gradient step, it would be a reasonable option to look at the reference models' recent discriminative confidences on the previous mini-batch. Importantly, while this method captures \cite{le2025multi}'s discriminative confidence, it does not overfit nearly as much as weights computed on mini-batch $t-1$ are used on mini-batch $t$.

In experiments, we found that SWCW was prone to nan-ing gradients on \texttt{UltraFeedback}, and thus we introduce a one-hot encoded variant, denoted SWCW-OH, that sets $\alpha_{\text{SWCW-OH}, k}^{(t)} = 1$ if and only if $\alpha_{\text{SWCW}, k}^{(t)} = \max_{j \in [1:K]} \alpha_{\text{SWCW}, j}^{(t)}$, else $0$. One-hot encoding reduces to standard DPO per mini-batch.

\paragraph{Online Thompson Sampling Weighting (TSW)} We cast the choice of $\bdalpha$ as a $K$-armed bandit, using Thompson sampling to balance exploration and exploitation (see Appendix \ref{additional-literature-review}). To have closed-form posteriors, we restrict $\bdalpha$ to the one-hot-encoded basis vectors $\{ \bde_1, \dots, \bde_K\}$: at step $t$, we select one reference model $k$ and run MRPO with $\bdalpha = \bde_k$ (i.e., single-reference DPO towards reference $k$). Let $\pi_\theta^{(t)}$ be the fine-tuned policy after gradient step $t$. We define a binary reward $r_t(\bdalpha)$ recording whether \textit{stochastic validation accuracy} improved at time $t$, where $\tilde{\mathcal{D}}_{\text{val}}^{(t)}$ is a stochastic subsample from the full validation set. We use a subsample instead of the full set due to compute constraints:
\begin{align*}
    r_t(\bdalpha) = \bdone\left( \text{acc}\left( \pi_\theta^{(t)}; \tilde{\mathcal{D}}_{\text{val}}^{(t)}\right) > \text{acc}\left( \pi_\theta^{(t-1)}; \tilde{\mathcal{D}}_{\text{val}}^{(t)}\right) \right),
\end{align*}
Intuitively, if validation accuracy improves after pulling towards reference model $k$, then perhaps we should continue doing so with increased probability. Moreover, although we use only one one-hot-encoded basis vector $\bde_k$ at a time, the time-average of the selections over multiple mini-batches may effectively behave like a soft mixture: the empirical frequency of chosen arms would likely correspond to a soft weight vector on references \textit{without the basis set constraint.}

Following Thompson sampling for Bernoulli bandits \citep{russo2018tutorial}, for each arm $k$, we initialize and maintain a probability distribution $\theta_k \sim \text{Beta}(\alpha_k^{(t)}, \beta_k^{(t)})$, with $\alpha_k^{(0)} = \beta_k^{(0)} = 5$ or $10$ (denoted as the \textit{prior initialization value}, or PIV). Different choice of PIV dictate the initial rate of exploration vs. exploitation. The expected reward of arm $k$ at time $t$ is $\mu_k^{(t)} := \alpha_k^{(t)} / (\alpha_k^{(t)} + \beta_k^{(t)})$.

For each minibatch $t$, we will sample $\tilde{\theta}_k^{(t)} \sim \text{Beta}(\alpha_k^{(t)}, \beta_k^{(t)})$ for each arm $k$ and choose the arm $k^* = \argmax_k \tilde{\theta}_k^{(t)}$. Then, we run MRPO with $\bdalpha = \bde_{k^*}$, compute $r_t(\bdalpha)$, and update $(\alpha_{k^*}^{(t+1)}, \beta_{k^*}^{(t+1)})$:
\begin{align*}
    \text{If} \; r_t(\bdalpha) = 1, \; \text{then} \; (\alpha_{k^*}^{(t+1)}, \beta_{k^*}^{(t+1)}) = (\alpha_{k^*}^{(t)} + 1, \beta_{k^*}^{(t)}), \; \text{else} \; (\alpha_{k^*}^{(t+1)}, \beta_{k^*}^{(t+1)}) = (\alpha_{k^*}^{(t)}, \beta_{k^*}^{(t)} + 1).
\end{align*}
Because the $\bdalpha$ are one-hot encoded, TSW reduces to DPO per mini-batch, just as in SWCW-OH.

\subsection{Experimental Setup}

\paragraph{Models} We will use \href{https://huggingface.co/Qwen/Qwen2.5-0.5B-Instruct}{Qwen2.5-0.5B-Instruct} as our base policy model, chosen for its instruction-tuned performance and low compute requirements. Inspired partly by \cite{le2025multi}, we will use \href{https://huggingface.co/meta-llama/Llama-3.1-8B-Instruct}{Llama-3.1-8B-Instruct}, \href{https://huggingface.co/mistralai/Mistral-7B-Instruct-v0.3}{Mistral-7B-Instructv0.3}, \href{https://huggingface.co/Qwen/Qwen2.5-0.5B-Instruct}{Qwen2.5-0.5B-Instruct}, \href{https://huggingface.co/Qwen/Qwen2.5-1.5B-Instruct}{Qwen2.5-1.5B-Instruct}, \href{https://huggingface.co/Qwen/Qwen3-4B-Instruct-2507}{Qwen3-4B-Instruct}, \href{https://huggingface.co/01-ai/Yi-1.5-9B-Chat}{Yi-1.5-9B-Chat}, and \href{https://huggingface.co/microsoft/Phi-3-medium-128k-instruct}{Phi-3-Medium-128K-Instruct} (with 14B parameters) as our $K=7$ reference models. For all experiments, we will use all $K=7$ reference models.
\vspace{-0.25cm}

\paragraph{Datasets} We will work with the following public preference datasets (i.e., $\{ (\bdx_i, y^+_i, y^-_i)\}$ pairs): \href{https://huggingface.co/datasets/HuggingFaceH4/ultrafeedback_binarized}{UltraFeedback (Binarized)} and \href{https://huggingface.co/datasets/RLHFlow/PKU-SafeRLHF-30K-standard}{PKU-safeRLHF}, denoted as ``UltraFeedback" and ``SafeRLHF" for brevity. These datasets are large, so we subsampled 5K/1K/1K examples each for our train/val/test splits, shuffled over $3$ seeds for UltraFeedback and $5$ seeds for SafeRLHF. 

\vspace{-0.25cm}

\paragraph{Experimental Details} For each (subsampled) dataset and proposed method, we will run 1 epoch of training with batch-size $B=50$ for SafeRLHF and $B=25$ for UltraFeedback\footnote{These different choices were due to UltraFeedback having much longer prompt and reply token-lengths on average than SafeRLHF. The choice of $3$ seeds for UltraFeedback vs. $5$ for SafeRLHF was for similar reasons.}, replicated over the aforementioned seeds. Finally, we will include (a) standard DPO on each of the the $K=7$ reference models; and (b) MRPO/MDPO using per-example weighting (i.e., ``Original") as baselines.

\vspace{-0.25cm}

\paragraph{Evaluation Metrics} Our evaluation metric is length-normalized test classification accuracy after training and over gradient timestep $t$ (see Eqn. \ref{eq:val-set-acc}, generalized to the test set). All results will be averaged over seeds with standard errors and $p$-values as appropriate.

Pleass see full details on our experiments in Appendix \ref{additional-experimental-details}.

\begin{comment}
The pre-model DPO loss formula is 
\begin{equation*}
     \mathbb{E}_{(x, y^+, y^-) \sim \mathcal{D}} \left[ - \log \sigma \left( \beta \left( \log \frac{\pi_\theta(y^+|x)}{\pi_{\text{ref}}(y^+|x)} - \log \frac{\pi_\theta(y^-|x)}{\pi_{\text{ref}}(y^-|x)} \right) \right) \right]
\end{equation*}
where $\pi_\theta$ refers to \href{https://huggingface.co/Qwen/Qwen2.5-0.5B-Instruct}{Qwen2.5-0.5B-Instruct} and $\pi_{\text{ref}}$ to any other model from the $K=7$ reference models we use.
\end{comment}

\vspace{-0.25cm}

\section{Results and Discussion}
\vspace{-0.25cm}

\subsection{Comparing Overall Performances}
\label{subsection:overall}
\vspace{-0.25cm}

\begin{figure}
    \centering
    \includegraphics[width=1.0\linewidth]{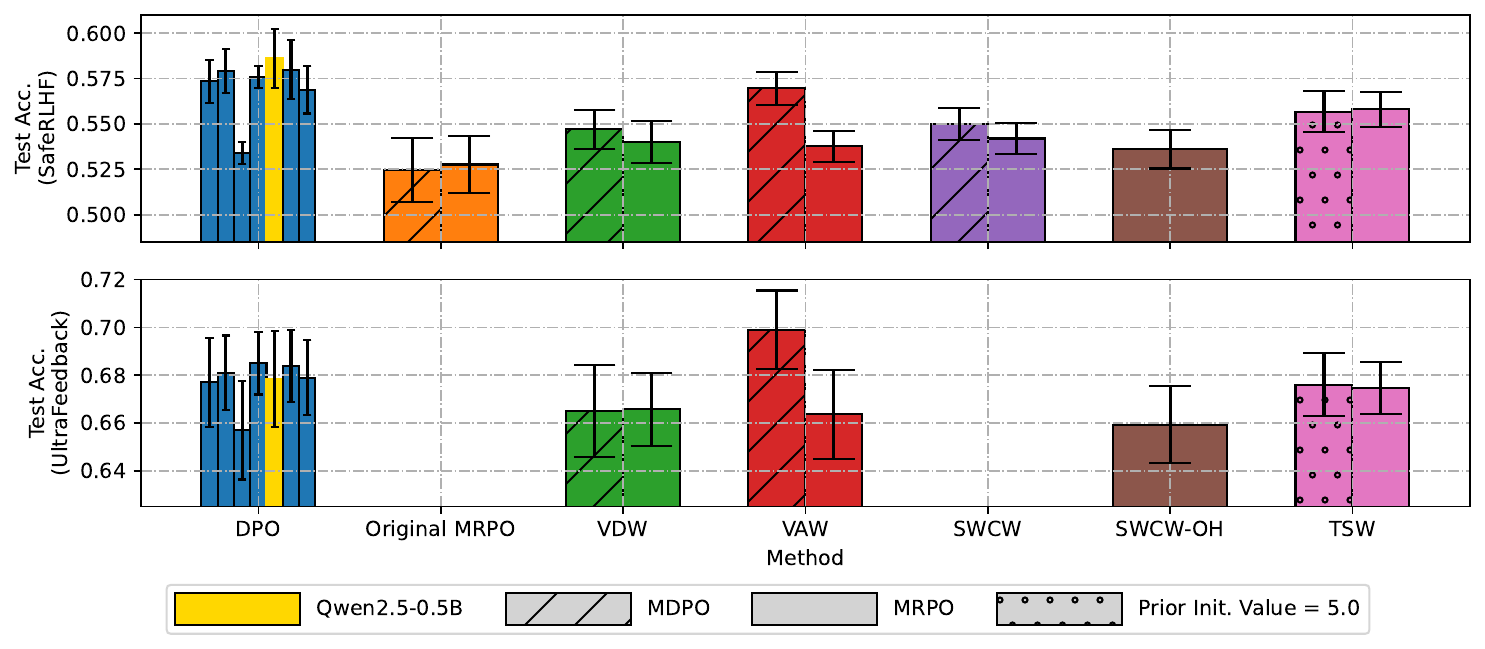}
    \vspace{-0.6cm}
    \caption{\footnotesize Test accuracies of (single-reference) DPO, original MRPO/MDPO, and our five proposed statistically-principled methods on SafeRLHF (top) and UltraFeedback (bottom), averaged across 5 and 3 randomly-seeded subsampled datasets, respectively, with standard errors. Diagonal hashes represent MDPO as opposed to MRPO, and polka-dots represent $\text{PIV}=5.0$ as opposed to $10.0$. Original MRPO/MDPO and SWCW are missing for UltraFeedback because of gradients invariably nan-ing out by mini-batch $5$. From left to right, each blue bar for DPO represents fine-tuning Qwen2.5-0.5B on the following reference models: Yi-1.5-9B, Llama-3.1-8B, Phi-3-Medium-128k, Mistral-7B, Qwen2.5-0.5B (marked in yellow), Qwen2.5-1.5B, and Qwen3-4B. \normalsize}
    \label{fig1:overall-performances}
\end{figure}

\paragraph{Comparison to Original MDPO/MRPO} From Figure \ref{fig1:overall-performances}, we observe that \textit{our five statistically-principled methods (VDW through TSW) uniformly outperform \citet{le2025multi}'s original MRPO and MDPO methods on mean test accuracy across $5$ seeds on SafeRLHF.} On UltraFeedback, both the original MRPO/MDPO methods and our SWCW method failed early --- invariably encountering \texttt{nan} gradients by mini-batch $5$ --- and thus underperformed relative to the other methods. Overall, our arguments about \citet{le2025multi}'s original MRPO scheme being suboptimal compared to more statistically-principled methods appears to ring true.

\begin{figure}
    \centering
    \includegraphics[width=1.0\linewidth]{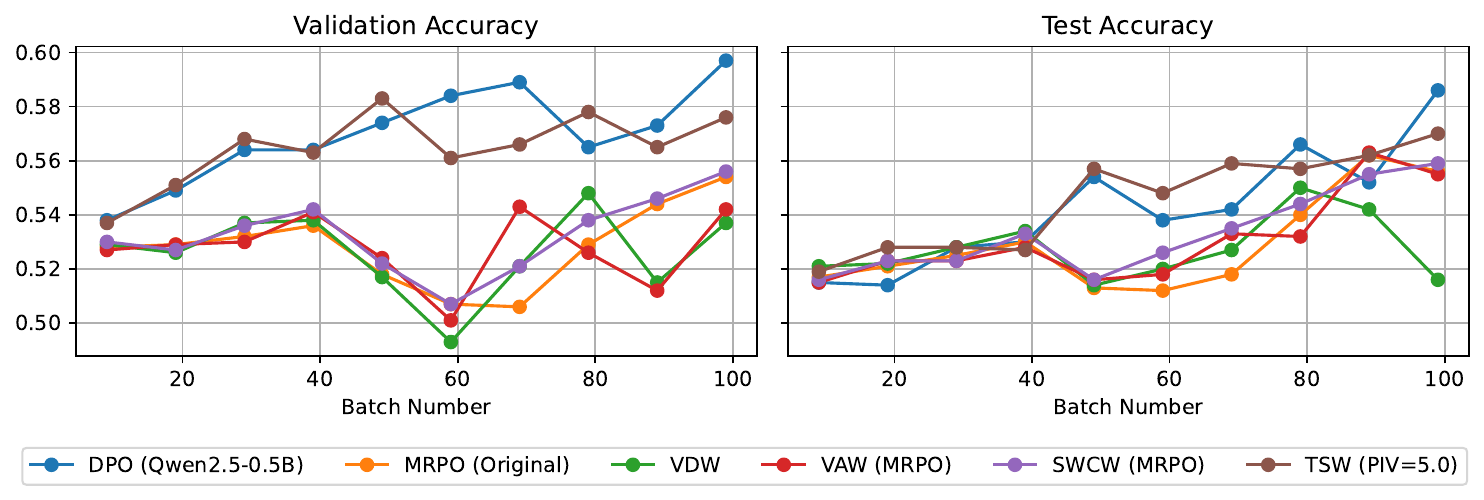}
    \vspace{-0.6cm}
    \caption{\footnotesize Validation (left) and test (right) accuracies over mini-batch of representative methods on SafeRLHF (seed 2). Such accuracy metrics are \textit{unstable across mini-batch and far from monotonic.}}
    \label{fig2:instability-of-accuracies}
\end{figure}

\vspace{-0.25cm}

\paragraph{Comparison to Single DPO} More thought-provokingly, \textit{DPO with a single reference model (for any one of 6 out of 7 choices of reference model) outperformed all other methods} on SafeRLHF. On UltraFeedback, only one of our statistically-principled methods, VAW (MDPO) is able to outperform the single DPO baselines on mean test accuracy, though the size of the error bars indicate modest statistical significance. \textit{Otherwise, single reference DPO on any one of 6 out of 7 choices of reference model outperformed all of our proposed methods on UltraFeedback}. Moreover, from Figure \ref{fig2:instability-of-accuracies}, we observe that single-reference DPO exhibits more stabler validation and test accuracies over mini-batches than all other methods: they decrease on only $1$ or $2$ mini-batches, compared to $4$ or even $5$ times for the other methods, out of only $10$ batches where said metrics were computed. One explanation is that the multiple KL terms of MRPO/MDPO-type approaches may induce competing and contradicting contributions to the model gradient, further obfuscating the fine-tuning signal.

\vspace{-0.25cm}

\paragraph{On MRPO vs. MDPO} From Figure \ref{fig1:overall-performances}, we observe that MDPO is effectively strictly preferable to MRPO across all methods, with the miniscule reversals on Original (SafeRLHF) and VDW (UltraFeedback) devoid of statistical significance by the size of the error bars. This presents a differing perspective from \citet{le2025multi}, though we do note that both MRPO and MDPO are still vulnerable to numerical instability, as shown on UltraFeedback.

\paragraph{On Numerical Instability}
Both the Original and VAW methods when paired with either the MDPO or MRPO loss functions fail to train on the UltraFeedback dataset, consistently resulting in NaNs. This behavior persists even after substantially reducing the learning rate and applying gradient-norm clipping. In contrast, the same methods train without issue on the SafeRLHF dataset. A plausible explanation is that UltraFeedback contains significantly longer responses, whereas SafeRLHF examples have much shorter sequence lengths. The increased token length in UltraFeedback thus likely amplifies instability in the optimization dynamics (despite our stable implementations), leading to numerical divergence under MDPO/MRPO. Further details on instability are deferred to Appendix \ref{regarding-numerical-stability}.

\vspace{-0.25cm}

\paragraph{Main Takeaways for General Preference Fine-Tuning} First, the \textit{traditional single-reference DPO may be preferable to multi-reference approaches}, regardless of how one weights the reference models. Second, \textit{the choice of reference model for said single-reference DPO is not of significant consequence.} Instead of hoping to distill some positive attributes of the reference model, we should view the reference model simply as a check on grammatical coherence: the vast majority of the fine-tuning signal comes from the preference dataset itself. One analogy for shedding insight on the above is the classical Ridge regression, with the following suggestively-rewritten objective function:
\begin{align}
    \beta^*_{\text{ridge}} = \argmin_{\beta} \| \bdy - X \beta \|_2^2 + \lambda \| \beta - \beta^*_{\text{ref}} \|_2^2, \quad \text{with } \beta^*_{\text{ref}} = \bdzero.
    \label{eqn:ridge}
\end{align}
In Ridge regression, the coefficients $\beta^*_{\text{ridge}}$ are \textit{primarily driven by the data $(X, \bdy)$ itself}, with the regularization term primarily to \textit{regularize} the optimization problem away from (near)-singularity and prevent $\beta^*_{\text{ridge}}$ from containing exorbitantly-large entries. No one believes that $\beta^*_{\text{ref}} = \bdzero$ is a promising solution with data/task-specific features worth ``distilling" towards. Indeed, in the classical literature, one rarely encounters the following objective, where the $\beta^*_{\text{ref},k}$ are apriori promising solutions:
\begin{align}
    \beta^* = \argmin_{\beta} \| \bdy - X \beta \|_2^2 + \sum_{k=1}^K \lambda_k \| \beta - \beta^*_{\text{ref},k} \|_2^2.
    \label{eq:crazy-ridge}
\end{align}
Rather, the experimentation around Ridge is primarily around selecting the best regularization strength $\lambda$ via cross-validation for Equation \ref{eqn:ridge}, as opposed to devising methods for selecting $(\lambda_1, \dots, \lambda_K)$ in Equation \ref{eq:crazy-ridge}. Analogously, for general preference fine-tuning, our experiments suggest that similar instincts may apply here: (a) \textit{the reference model should only be viewed as a check on grammatical coherence and regularity; (b) The majority of the fine-tuning signal is from our $(\bdx, y^+, y^-)$ preference pairs; and (c) a single reference will more than suffice.}

\subsection{Among Our Proposed Methods}
\vspace{-0.25cm}
On both datasets, VAW (MDPO) was the strongest of our proposed statistically-principled methods, with TSW in second place. We will defer discussion on TSW to the next subsection.

\begin{figure}
    \centering
    \includegraphics[width=1.0\linewidth]{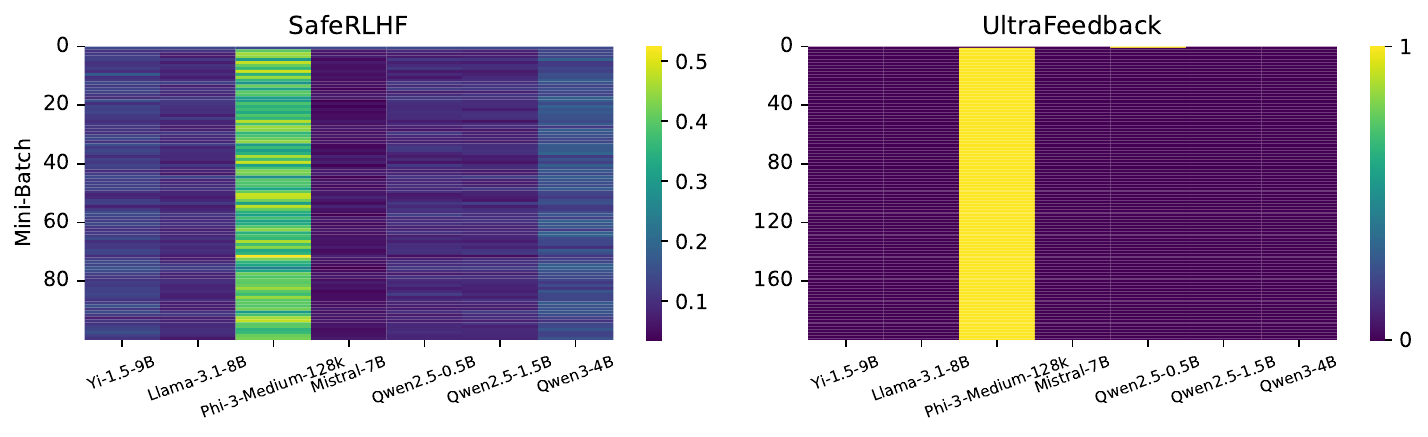}
    \vspace{-0.6cm}
    \caption{\footnotesize Representative $\bdalpha \in \Delta^6$ values per reference model over mini-batch from SWCW on SafeRLHF (left) and SWCW-OH on UltraFeedback (right). Results are shown for seed 2, but trends are consistent over all seeds.}
    \label{fig3:online-1-alpha-values}
\end{figure}

VDW and SWCW's poor performance can likely be attributed to the fact that on both datasets, Microsoft's Phi-3-Medium-128k was the most overconfident and incorrect, yielding a corresponding $\alpha_k$ value that dominated all other reference models' (see Figure \ref{fig3:online-1-alpha-values}). Because both VDW and SWCW weight references proportional to discriminative confidence (see Equations \ref{eq:VDW} and \ref{eq:SWCW}), these methods placed the bulk of their weight on Phi-3, resulting in (poor) performance roughly similar to single-reference DPO to Phi-3. To finish, SWCW-OH demonstrates similar performance to VDW on UltraFeedback. This is not surprising as both construct $\alpha_k$'s based on discriminative confidence. SWCW-OH's slightly worse performance than SWCW on SafeRLHF can likely be explained by its one-hot encoding nature, which we will discuss more in Section \ref{tsw-analysis}.

\begin{table}[]
    \centering
    \caption{\footnotesize Kendall's-$\tau$ rank correlation coefficients $\tau$ between the $\bdalpha \in \Delta^6$ weights per reference model used by VAW and VDW; and either (a) each reference models' test accuracy (e.g., Llama-3.1-8B's own performance on UltraFeedback), denoted ``Ref. Acc."; or (b) the final test accuracies of Qwen2.5-0.5B fine-tuned via DPO on each reference model individually, denoted ``Single DPO Acc," averaged across seeds with finite-sample-exact $p$-values. A $\tau$ value of $1$ indicates perfect rank correlation, a value near $0$ indicates no rank correlation (i.e., random guessing), and a value of $-1$ implies worse-than-guessing alignment. Correlations between the references' test accuracies and the final test accuracies of Qwen2.5-0.5B fine-tuned via DPO on said references are shown in the last column (averaged across seeds, with standard errors).
    \normalsize}
    \begin{adjustbox}{width=1\textwidth}
    \begin{tabular}{llrrr}
    \toprule
     \textbf{Dataset} & \textbf{Method} & \makecell{\textbf{Kendall-$\tau$}\\($\bdalpha$, Ref. Acc.)} & \makecell{\textbf{Kendall-$\tau$}\\($\bdalpha$, Single DPO Acc.)} & \makecell{\textbf{Correlation}\\(Ref. Acc., Single DPO Acc.)} \\
    \midrule
    \multirow[t]{2}{*}{SafeRLHF} & VAW & 0.481 ($p<0.001$) & 0.017 ($p=0.443$) & 0.254 $\pm$ 0.101 \\
     & VDW & -0.109 ($p=0.788$) & -0.457 ($p=0.999$) & \\
    \cr
    \multirow[t]{2}{*}{UltraFeedback} & VAW & 0.396 ($p=0.011$) & 0.174 ($p=0.150$) & 0.058 $\pm$ 0.312 \\
     & VDW & -0.015 ($p=0.508$) & -0.301 ($p=0.943$) &
    \cr
    \bottomrule
    \end{tabular}
    \end{adjustbox}
    \label{tab:offline-methods-alpha-kendalls}
\end{table}

VAW’s superior performance is likely due to two related effects: (a) on SafeRLHF, reference models with higher test accuracy tend to induce higher final test accuracy for Qwen2.5-0.5B under single-reference DPO (see ``Correlation" in Table \ref{tab:offline-methods-alpha-kendalls}); and (b) VAW assigns larger $\boldsymbol{\alpha}$ weights to those same higher-accuracy reference models, as its $\boldsymbol{\alpha}$ values are moderately to strongly rank-correlated with the reference models’ test accuracies (as measured via Kendall's-$\tau$ rank correlation)\footnote{We use Kendall's-$\tau$ rank correlation in Table \ref{tab:offline-methods-alpha-kendalls} to assess the relationships between $\bdalpha$ and accuracies because $\bdalpha$ values, depending on method, can be much more dispersed than accuracies. We still use standard correlation to quantify relationships between accuracy metrics, as they are on the same scale with comparable dispersion.}. While the (mean) correlation between reference models' test accuracy and single-reference DPO final test accuracy is not nearly as high on UltraFeedback, the standard error across seeds ($0.312$) is massive, suggesting that, on some seeds, such a (rank)-correlation is still quite substantial. From a complementary perspective, our reference models, for the most part (c.f. Phi-3), exhibited generally comparable validation accuracies, which resulted in VAW having weights that were relatively closer to uniform than any of the other methods. It is also possible that such a uniform weighting injects a corresponding variance reduction in gradient updates, though this is conjecture for now.

In passing, from Table \ref{tab:offline-methods-alpha-kendalls}, we further confirm that VDW's assigned $\bdalpha$ values per reference are not rank-correlated (and sometimes even anti-rank-correlated, due to Phi-3) with single-reference DPO final test accuracies, further explaining VDW's low performance.

\subsection{On Thompson Sampling (TSW)}
\label{tsw-analysis}
\vspace{-0.25cm}

\begin{figure}
    \centering
    \includegraphics[width=1.0\linewidth]{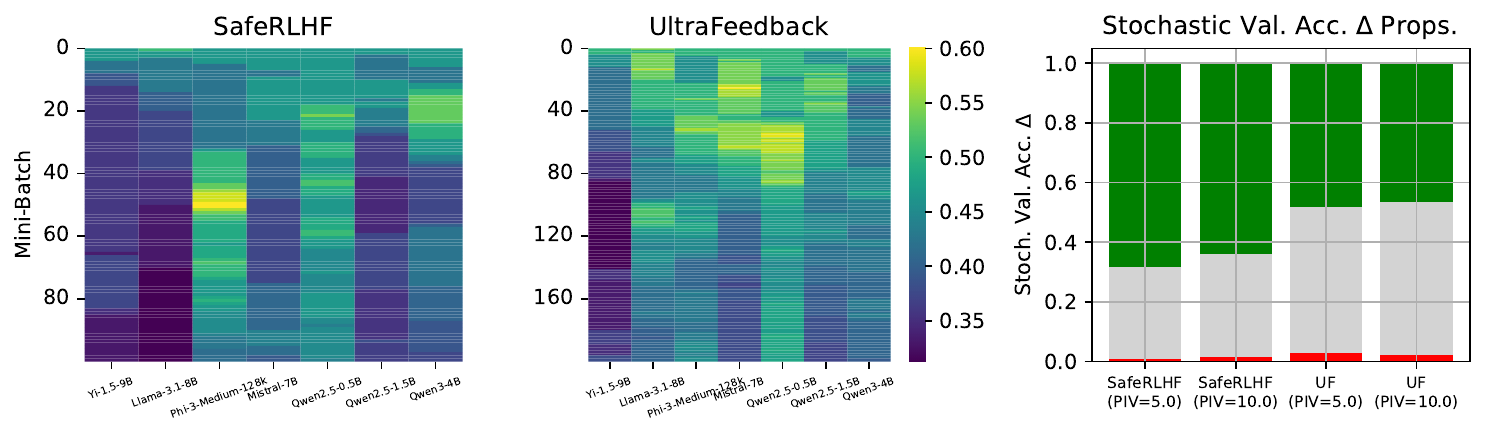}
    \vspace{-0.6cm}
    \caption{\footnotesize Representative $\bdmu \in \R^7$ estimated mean rewards per reference model over mini-batch from TSW ($\text{PIV}=5.0$) on SafeRLHF (left) and UltraFeedback (center), seed 0. The rightmost subplot shows the proportions of mini-batches per dataset and PIV setting where $\Delta$, the change in stochastic validation set accuracy pre- and post- gradient step, was negative (red), zero (grey), or positive (green), averaged over all seeds. \normalsize}
    \label{fig4:online-2-mu-values}
\end{figure}

Apriori, we expected the online TSW to perform the best out of our proposed statistically-principled methods, as we predicted that it would statistically-principledly learn the optimal reference model to align towards. Indeed, from Figure \ref{fig4:online-2-mu-values} (Left + Center), we observe that the estimated mean rewards per reference model do change significantly over mini-batch, suggesting that the corresponding online $\bdalpha$'s are not effectively-frozen to a suboptimal arrangement as was in SWCW, our other online method. We do note from Figure \ref{fig1:overall-performances} that TSW's test accuracy was consistently higher than the worst-performing single-DPO (i.e., Phi-3), a bar that, of our proposed methods, only VAW was also able to meet.

However, from the rightmost panel of Figure \ref{fig4:online-2-mu-values}, we note that our reward function $r_t(\bdalpha)$ is inherently very noisy: in nearly 30-40\%+ of mini-batches, stochastic validation accuracy shows zero change (positive or negative) after a DPO gradient step with the chosen reference. This is likely because LLMs have large parameter counts, and taking one gradient step after a mini-batch of 25 or 50 preference pairs may not significantly influence the LLM's overall performance on the fine-tuning task. Indeed, from Table \ref{tab:online-2-mus-kendalls}, we observe that the final estimated mean rewards per reference model exhibit little rank-correlation to single-reference DPO test accuracy, suggesting that the reward environment was too noisy for meaningful acquisition of the optimal reference model. We 

Finally, TSW’s one-hot encoded $\boldsymbol{\alpha}$'s also introduce additional gradient instability. In the non-one-hot offline multi-reference methods discussed in Section~\ref{subsection:overall}, the conflicting KL-based contributions still produce a relatively consistent (though suboptimal) KL-pulling direction across mini-batches, so the obfuscation of the fine-tuning signal is at least stable. In contrast, one-hot encoded $\boldsymbol{\alpha}$’s cause the active reference to change from mini-batch to mini-batch, inducing highly variable KL-pulls and further amplifying the obfuscation of the true fine-tuning signal. Intuitively, a gradient direction that works well when fine-tuning toward Mistral-7B is not necessarily appropriate once the next mini-batch instead pulls toward Yi-1.5-9B. Moreover, Thompson sampling assumes (approximately) stationary reward distributions across mini-batches, an assumption that appears to be violated here.

\begin{table}
    \centering
    \caption{\footnotesize Kendall's-$\tau$ rank correlation coefficients between the final $\bdmu \in \R^7$ estimated mean rewards per reference model used by TSW; and either (a) the final test accuracies of Qwen2.5-0.5B fine-tuned via DPO on each reference model individually, denoted ``Single DPO Acc."; or (b) each reference models' test accuracy, denoted ``Ref. Acc.", averaged across seeds with finite-sample-exact $p$-values. See Table \ref{tab:offline-methods-alpha-kendalls} for further explanation.}
    \begin{adjustbox}{width=0.67\textwidth}
    \begin{tabular}{llrr}
    \toprule
     \textbf{Dataset} & \textbf{PIV} & \makecell{\textbf{Kendall-$\tau$}\\($\bdmu$, Single DPO Acc.)} & \makecell{\textbf{Kendall-$\tau$}\\($\bdmu$, Ref. Acc.)} \\
    \midrule
    \multirow[t]{2}{*}{SafeRLHF} & $5.0$ & -0.018 ($p=0.550$) & 0.000 ($p=0.497$)\\
     & $10.0$ & 0.278 ($p=0.021$) & -0.098 ($p=0.746$) \\
    \cr
    \multirow[t]{2}{*}{UltraFeedback} & $5.0$ & -0.142 ($p=0.757$) & -0.047 ($p=0.584$)\\
     & $10.0$ & -0.142 ($p=0.757$) & 0.333 ($p=0.028$)
    \cr
    \bottomrule
    \end{tabular}
    \end{adjustbox}
    \label{tab:online-2-mus-kendalls}
\end{table}
\vspace{-0.25cm}

\section{Limitations and Future Work}
\vspace{-0.25cm}
Due to limited compute, our results were only over $3$ seeds for UltraFeedback and $5$ seeds for SafeRLHF. Our conclusions should be verified with many more seeds in the future. Relatedly, we did not do full-scale hyperparameter tuning (e.g., regularization strength $\beta$ and learning rate). Our base model Qwen2.5-0.5B-Instruct is also on the smaller side and training dynamics could change if we had a more powerful base model. In addition, our reference models generally had comparable performance on UltraFeedback and SafeRLHF: our findings could potentially change if there were larger differences between our reference models. On a similar note, we only did $1$ epoch of training per method using just 5K training examples and proportionate validation sets: future work could look into multiple epochs and much larger training sets. Our findings regarding accuracy stability over mini-batches could change with multiple epochs. Moreover, we do note that production-level utilities like \texttt{DPOTrainer} have additional engineering optimizations under-the-hood that were not readily available to us for our specific experimental setups (see Appendix \ref{additional-experimental-details}).

\vspace{-0.25cm}
\section{Conclusion}
\vspace{-0.25cm}

In this paper, we proposed four statistically-principled strategies for weighting reference models under the MRPO/MDPO frameworks --- two offline methods based on validation sets, one online method based on a sliding-window, and one online method using Thompson sampling. While all four of our proposed strategies considerably outperform the current MRPO weighting strategy proposed by \citet{le2025multi}, single-reference DPO --- almost regardless of the choice of reference model --- consistently outperformed all evaluated multiple-reference approaches (except VAW on UltraFeedback). Our findings call into question the practical utility of using multiple-reference approaches for preference fine-tuning. Indeed, based on our results, we postulate that the role of the reference model is mainly to serve as a check on grammatical coherence and regularity, as opposed to a target with task-specific qualities worth distilling into our policy model. The majority of the fine-tuning signal comes from the preference dataset itself. Perhaps one reference model is enough.

%

%

%

%

%

%

%

%

%

%

%
\begin{comment}
%
%
%
\end{comment}

\newpage
\bibliography{references}
\bibliographystyle{unsrtnat}

\newpage

\appendix

\section{Appendix \ref{additional-literature-review}}
\label{additional-literature-review}

\subsection{Additional Details on RLHF}
\cite{ziegler2019fine} proposes finetuning LLMs using human preference signals. Given a dataset of prompts \(x \in X\) and a human choice \(b \) from a finite set of choices \(y_1,..,y_N \in Y\), they build a reward model $r : X \times Y \to \mathbb{R}$ by copying a pretrained LLM $\rho$ and randomly initializing its final layer. Then, they train $r$ using the loss
\[\mathrm{loss}(r)
= \mathbb{E}_{(x, \{y_i\}, b) \sim S}
\left[
  \log \frac{e^{r(x, y_b)}}{\sum_i e^{r(x, y_i)}}
\right], \text{encouraging a higher reward on the chosen output.}\]
Starting from the reference model $\rho$, the policy $\pi$ is fine-tuned with Proximal Policy Optimization (PPO) \citep{schulman2017proximal} on the following reward:
$R(x, y) = r(x, y) - \beta \log \frac{\pi(y \mid x)}{\rho(y \mid x)}$, where the subtracted term reguarlizes $\pi$ towards $\rho$ with weight $\beta$. This is a standard RLHF pipeline: (a) supervised fine-tuning (SFT) to initiate a helpful $\rho$; (b) preference sampling and training a reward model; and (c) reinforcement learning optimization (here, PPO).

In SFT, we begin by fine-tuning a pre-trained language model on high-quality human-written data for downstream tasks such as dialogue or summarization. The fine-tuned model is denoted by $\pi^{\text{SFT}}$. The SFT model is then used to generate multiple responses $y_1, y_2 \sim \pi^{\text{SFT}}(y \mid x)$ for a given prompt $x$. Humans then express preferences between the generated outputs, denoted by $y^+ \succ y^-$, where $y^+$ and $y^-$ correspond to the preferred and non-preferred completions, respectively. These preferences are assumed to originate from an underlying reward model $r^*(y, x)$ which is not directly observable. A common model for human preferences is the Bradley-Terry (BT) model, which defines the probability that $y^+$ is preferred to $y^-$ given $x$ as:
\begin{equation}
    p^*(y^+ \succ y^- \mid x) = \frac{\exp(r^*(x, y^+))}{\exp(r^*(x, y^+)) + \exp(r^*(x, y^-))} = \sigma(r^*(x, y^+)-r^*(x, y^-))
    \label{eq:1}
\end{equation}
Given a dataset of comparisons $\mathcal{D} = \{(x_i, y^+_i, y^-_i)\}_{i=1}^N$ sampled from $p^*$, the reward model $r_\phi(x, y)$ can be learned by minimizing the negative log-likelihood:
\[ -\mathbb{E}_{(x, y^+, y^-) \sim \mathcal{D}} \left[ \log \sigma\big(r_\phi(x, y^+) - r_\phi(x, y^-)\big) \right]\]
where $\sigma$ denotes the sigmoid function. In practice, $r_\phi(x, y)$ is often obtained by extending the SFT model $\pi^{\text{SFT}}$ with an additional randomly initialized linear head on top of the final layer that outputs a scalar value as described in \cite{stiennon2020learning}. During the RL phase, the learned reward function provides feedback for updating the policy. Following~\cite{stiennon2020learning}, the optimization objective becomes:
\begin{equation}
    \max_{\pi_\theta} \; \mathbb{E}_{x \sim \mathcal{D}, y \sim \pi_\theta(y \mid x)} \big[r_\phi(x, y)\big] - \beta D_{\text{KL}}\!\left(\pi_\theta(y \mid x) \,\|\, \pi_{\text{ref}}(y \mid x)\right)
    \label{eq:2}
\end{equation}
where $\pi_{\text{ref}}$ is typically initialized to $\pi^{\text{SFT}}$, and $\beta$ controls the strength of deviation from this reference policy. This optimization is performed with RL algorithms such as PPO \citep{schulman2017proximal}.  

\subsection{Length-Normalized Direct Preference Optimization (LN-DPO)}

LN-DPO is introduced by \cite{ahrabian2025practical} as a length-normalized variant of Direct Preference Optimization (DPO) that improves stability during finetuning while maintaining alignment performance. The motivation arises from a known issue in DPO: the implicit reward term
\[
\log \frac{\pi_\theta(y \mid x)}{\pi_{\mathrm{ref}}(y \mid x)}
\]
tends to increase with the response length $|y|$, which often causes DPO to favor unnecessarily long and lower-quality responses. To mitigate this, LN-DPO normalizes the reward by the response length. Its optimization objective is
\begin{equation}
\mathbb{E}_{(x, y^+, y^-) \sim \mathcal{D}}
\left[
  -\log \sigma\!\left(
    \frac{\beta}{|y^+|}
      \log\frac{\pi_\theta(y^+ \mid x)}{\pi_{\mathrm{ref}}(y^+ \mid x)}
    -
    \frac{\beta}{|y^-|}
      \log\frac{\pi_\theta(y^- \mid x)}{\pi_{\mathrm{ref}}(y^- \mid x)}
  \right)
\right],
\label{eq:ln-dpo}
\end{equation}

which rescales each preference comparison by the lengths $|y^+|$ and $|y^-|$. This normalization removes the artificial advantage that long responses receive under vanilla DPO, enabling the model to focus more on preference pairs with strong evidence of quality differences. As noted in the original paper, this corresponds to introducing an adaptive margin, where pairs that the reference model already distinguishes clearly receive a larger effective influence. Empirically, LN-DPO produces more concise responses, shows improved robustness across hyperparameters, and more importantly is more stable during finetuning.

\subsection{KL-Regularized RLHF with Multiple Reference Models}
Standard RLHF and DPO typically regularize against a single reference model, limiting their ability to exploit the diversity of the many pre-trained LLMs now available. \cite{aminian2025theoretical} address this limitation by deriving the first exact solution to the multi-reference RLHF problem. Under reverse KL regularization, they study the objective

\begin{equation}
\max_{\pi_\theta} \; 
\mathbb{E}_{x \sim \rho, \, y \sim \pi_\theta(\cdot|x)} [r(x, y)]
- \beta \left( 
\sum_{k=1}^{K} \alpha_k 
D_{\mathrm{KL}}(\pi_\theta \| \pi_{\text{ref}}^k)
\right),
\label{eq:mrpo-rlhf}
\end{equation}
where \( \alpha_k \) are non-negative weighting coefficients satisfying \( \sum_{k=1}^K \alpha_k = 1 \)
\begin{comment}
\begin{equation}
\max_{\pi}\;
\mathbb{E}_{Y\sim\pi(\cdot|x)}\!\left[r_{\theta^\star}(x,Y)\right]
\;-\;
\beta\sum_{i=1}^{K} \alpha_i\,
D_{\mathrm{KL}}\!\big(\pi(\cdot|x)\,\Vert\,\pi_{\mathrm{ref},i}(\cdot|x)\big)
\label{eq:mrpo-rlhf}
\end{equation}

where the mixture weights satisfy $\alpha_i\in(0,1)$ and $\sum_{i=1}^K \alpha_i = 1$.  
\end{comment}

In contrast to MRPO, which optimizes only a lower-bound surrogate of this objective, \cite{aminian2025theoretical} show that the optimal policy takes the exponential-family form
\[
\pi_{\theta^\star}(y\mid x)
\;\propto\;
\hat{\pi}_{\alpha,\mathrm{ref}}(y\mid x)\,
\exp\!\left(\frac{1}{\beta}\, r_{\theta^\star}(x,y)\right)
\]
where the effective reference model is the generalized geometric mixture
\[
\hat{\pi}_{\alpha,\mathrm{ref}}(y\mid x)
=
\frac{\prod_{i=1}^K \pi_{\mathrm{ref},i}(y|x)^{\alpha_i}}
     {\sum_{y'\in\mathcal{Y}}
     \prod_{i=1}^K \pi_{\mathrm{ref},i}(y'|x)^{\alpha_i}}
\]
This mixture differs fundamentally from the harmonic mean $\tilde{\pi}_{\mathrm{ref}}$ used in MRPO, which arises from optimizing a tractable lower bound rather than the exact RLHF objective.
Substituting $\hat{\pi}_{\alpha,\mathrm{ref}}$ into the objective yields the equivalent problem
\[
\max_{\pi}\;
\mathbb{E}_{Y\sim\pi(\cdot|x)}[r_{\theta^\star}(x,Y)]
\;-\;
\beta\, D_{\mathrm{KL}}\!\big(
\pi(\cdot|x)\,\Vert\,\hat{\pi}_{\alpha,\mathrm{ref}}(\cdot|x)
\big)
\]
Thus, multi-reference RLHF is equivalent to standard reverse KL–regularized RLHF with a single geometric reference policy. 

\begin{comment}
By taking the reward estimator to be
\[
\hat{\theta}
= \arg\max_{\theta}
\mathcal{L}_R(\theta,\mathcal{D}),
\quad \text{where} \quad
\mathcal{L}_R(\theta,\mathcal{D})
= \frac{1}{n}\sum_{i=1}^n
\log\sigma\!\left(r_{\theta}(x_i, y_i^{w})
 - r_{\theta}(x_i, y_i^{\ell})\right)
\]
and defining the corresponding policy
\[
\pi_{\hat{\theta}}(y\mid x)
\;\propto\;
\hat{\pi}_{\alpha,\mathrm{ref}}(y\mid x)\,
\exp\!\left(\frac{1}{\beta}\, r_{\hat{\theta}}(x,y)\right)
\]
\cite{aminian2025theoretical} show an $O(1/n)$ upper bound on the KL-regularized sub-optimality gap under standard assumptions of bounded reward, finite reward class, and local reverse KL-ball coverage with respect to $\hat{\pi}_{\alpha,\mathrm{ref}}$. 
\end{comment}

Similar to MRPO, we note that all empirical validations in \cite{aminian2025theoretical} are carried
out exclusively in the two-reference setting ($K=2$). Their experiments
implement only the $K=2$ case, and no experiments investigate whether the theoretical guarantees remain numerically stable when combining a larger collection of reference
models. Thus, while the theory is fully general in $K$, the practical behavior of multi-reference RLHF for $K>2$ remains untested
and is an open question for future empirical study.

\subsection{Multi-Direct Preference Optimization (Multi-DPO)}

A closely related approach to MRPO is Multi-DPO (MDPO), a naive extension of DPO to the multi-reference setting. Rather than modifying the underlying RLHF objective as in MRPO, MDPO simply forms a weighted sum of $K$ independent DPO losses, one per reference model. It minimizes the following loss
\[
-\mathbb{E}_{x, y_w, y_\ell \sim \mathcal{D}}
\left[
\sum_{k=1}^{K} \alpha_k
\log\sigma\!\left(
\beta \log\frac{\pi_\theta(y^+ \mid x)}{\pi_{\mathrm{ref}}^{k}(y^+ \mid x)}
-
\beta \log\frac{\pi_\theta(y^- \mid x)}{\pi_{\mathrm{ref}}^{k}(y^- \mid x)}
\right)
\right]
\]
This formulation treats each reference model independently and aggregates their DPO losses linearly. While simple to implement, MDPO does not arise from optimizing a KL-regularized RLHF formulation. Consequently, MDPO lacks the probabilistic consistency of a principled multi-reference method. As with MRPO, all reported experiments in \citet{le2025multi} evaluate MDPO only in the case $K=2$, leaving its behavior for larger \(K\) unexplored.

\subsection{Thompson Sampling for Bernoulli Multi-Armed Bandits}
\vspace{-0.25cm}
The Bernoulli multi-armed bandit setup consists of $K$ possible actions $a_k$: each action $k$ when played at time $t$ outputs a 0-1 reward $r_t \sim \text{Bern}(\theta_k)$, with $\theta_k \in [0, 1]$ unknown to the player. The goal of a bandit problem is to explore and exploit towards maximizing total reward over time: in other words, the agent sequentially takes actions, observes rewards, and tries to learn the optimal strategy (i.e., arm) \citep{russo2018tutorial}. Thompson Sampling, as explained in \cite{russo2018tutorial}, is a Bayesian algorithm for solving the multi-armed bandit setup by balancing exploration and exploitation. First, the agent begins by placing a prior probability distribution on each arm $k$'s reward success parameter $\theta_k$ to capture their beliefs about said arm, with \cite{russo2018tutorial} recommending the choice of $\theta_k \simIND \text{Beta}(\alpha_k, \beta_k)$ for closed-form Bayesian conjugacy. Then, at each round $t$, the agent (a) draws $\tilde{\theta}_k \simIND \text{Beta}(\alpha_k, \beta_k)$; (b) takes action $k^* = \argmax_k \tilde{\theta}_k$; and (c) observes reward $r_t \in \{ 0, 1 \}$. Having observed the reward, the agent updates her beliefs about $\theta_{k^*}$. Specifically, by Bayes' Rule and Beta-Bernoulli Conjugacy, $\theta_{k^*} \mid r_t \sim \text{Beta}(\alpha_{k^*} + r_t, \beta_{k^*} + (1 - r_t))$, with the posterior distributions of the unselected arms $k$ untouched. Compared to more greedy algorithms, Thompson Sampling ``explores to resolve uncertainty where there is a chance that resolution will help the agent identify the optimal action, but avoids probing where feedback would be not helpful" \cite{russo2018tutorial}. 
%

%

%
%
%
%
%
%
%
%
%
%
%
%
%
%
%
%
%

\begin{comment}
After observing $y_t$, the posterior distribution is updated using Bayes’ rule:
\[
p(\theta = u \mid x_t, y_t) = \frac{p(u) \, q_u(y_t \mid x_t)}{\sum_{u'} p(u') \, q_{u'}(y_t \mid x_t)}
\]
As will be seen later, we will use the following model where each arm $k \in \{1, \dots, K\}$ corresponds to a basis vector $e_k$ and the algorithm observes a binary reward
\[
r_t \in \{0,1\}
\]  
    Unlike the Bernoulli bandit, where each arm maintains an independent Beta posterior over its success probability $\theta_k$, here all arms are modeled jointly using Dirichlet distributed probability vector
\[
\theta = (\theta_1, \dots, \theta_K) \sim \mathrm{Dirichlet}(\alpha)
\]
where $\alpha = (\alpha_1, \dots, \alpha_K)$ are positive parameters, commonly denoted as pseudo-observations. 

The adaptation of Thompson Sampling to this case is that at each iteration $t$ we do the following :

\begin{enumerate}
    \item Draw a probability vector from the current posterior:
    \[
    \hat{\theta} \sim \mathrm{Dirichlet}(\alpha)
    \]
    \item Choose the arm with the highest sampled weight:
    \[
    x_t = \arg\max_{k \in \{1, \dots, K\}} \hat{\theta}_k
    \]
    \item Take action $x_t$ and observe binary reward $r_t$
    \item The Dirichlet posterior is updated using 
    \[
    \alpha_{x_t} \leftarrow \alpha_{x_t} + r_t
    \]
    where the selected arm is $x_t$. 
    
\end{enumerate}
\end{comment}

\section{Additional Details on Experiments}
\label{additional-experimental-details}

\subsection{Additional Details on Experiments}

\paragraph{Dataset Subsampling} Because of limited compute, we reduced the maximum context length of our models to $L=1024$. Accordingly, when subsampling from UltraFeedback and SafeRLHF to form our seeded train/val/test splits, we only sampled preference pairs that (a) had prompt token lengths greater than the $2.5$th percentile (to avoid empty/corrupted prompts); (b) had both chosen and rejected response token lengths greater than the $2.5$th percentile (to avoid empty, corrupted, or yes/no and single-number answers); and (c) had total prompt+reply length (for both chosen and rejected) less than $L=1024$ to avoid any trimming of outputs and preserve fidelity.

\paragraph{DPOTrainer (TRL) and Logit Precomputation} 
We do not employ the \texttt{DPOTrainer} implementation from the TRL library for two primary reasons. 
First, the current version of \texttt{DPOTrainer} does not support multi-reference preference optimization, which is a central requirement in our setting where each example may be evaluated against $K>1$ reference policies. Second, while \cite{aminian2025theoretical} propose an ensemble-based strategy for reference modeling that still uses \texttt{DPOTrainer}, such an approach is computationally prohibitive in our context as it requires constantly recomputing reference models' logits. To remain within our compute budget, we instead precomputed all $\pi_{\mathrm{ref}}$ quantities offline and use them directly during training. This design choice enables faithful multi-reference learning while avoiding the overhead of repeatedly evaluating a reference ensemble during optimization.

\paragraph{Hyperparameters} Following best practices in the literature, we used a DPO/MRPO/MDPO regularization strength of $\beta = 0.1$ and a learning rate of $1 \times 10^{-4}$ for all experiments. To alleviate numerical instability we did do small-scale trials with smaller learning rates and/or gradient clipping, but these attempts were not successful --- all of this is reflected in our main text figures. We did not have enough compute for full-scale hyperparameter sweeps.

\paragraph{Model Training} Because of limited compute, we used \texttt{bitsandbytes}, \texttt{accelerate}, and \texttt{peft} to fine-tune our base model Qwen2.5-0.5B using LoRA and 4-bit quantization. Please see our GitHub repo at \href{github.com/AymenEcharghaoui/intelligently-weighted-finetuning}{github.com/AymenEcharghaoui/intelligently-weighted-finetuning} for full details.

\section{Regarding Numerical Instability}
\label{regarding-numerical-stability}

In our own implementation, we computed the MRPO loss using a numerically stable
\texttt{logsumexp} formulation. Writing
\[
\ell_\theta^\pm
= \frac{1}{|y^\pm|}\log \pi_\theta(y^\pm \mid x)
\quad \text{and} \quad
\ell_{k}^\pm
= \frac{1}{|y^\pm|}\log \pi_{\mathrm{ref}}^{k}(y^\pm \mid x)
\]

The MRPO logit for a single
triple $(x,y^+,y^-)$ can be written as
\begin{align*}
z_{\text{MRPO}}
&= \beta\Big[
  \big(\ell_\theta^+ - \ell_\theta^-\big)
  - \log\!\Big(\sum_{k=1}^K \alpha_k e^{-\ell_{k}^+}\Big)
  + \log\!\Big(\sum_{k=1}^K \alpha_k e^{-\ell_{k}^-}\Big)
  \Big] \\
&= \beta\Big[
  \big(\ell_\theta^+ - \ell_\theta^-\big)
  + \big(L_{\mathrm{ref}}^- - L_{\mathrm{ref}}^+\big)
  \Big]
\end{align*}
where $L_{\mathrm{ref}}^\pm = -\log \sum_{k} \alpha_k e^{-\ell_k^\pm}$ is implemented via
\texttt{logsumexp}. Even though \texttt{logsumexp} prevents catastrophic overflow/underflow
inside each sum, the term $L_{\mathrm{ref}}^- - L_{\mathrm{ref}}^+$ is still numerically ill-conditioned. For uniformly distributed \(\alpha_k\), the terms \(L_{\mathrm{ref}}^\pm\) implement a soft minimum over the normalized reference log-probabilities:
\[
L_{\mathrm{ref}}^\pm \approx \min_{k} \ell_k^\pm
\]
As the sequence length increases, the normalized log-probabilities
$\ell_k^\pm$ concentrate around their per-token means. Consequently, the MRPO correction term $L_{\mathrm{ref}}^- - L_{\mathrm{ref}}^+$
is the difference of two quantities each behaving like a minimum over~$K$
concentrated values. This difference of soft-mins is highly sensitive to
tiny perturbations in the underlying $\ell_k^\pm$, leading to unstable logits
and saturated gradients.

MDPO also exhibits practical instability. Because the method forms a separate DPO
loss for each reference model and then aggregates them, the gradients coming from
different references often point in conflicting directions. Small differences in
how each reference scores $y^+$ versus $y^-$ can cause some MDPO terms to saturate
while others remain active, so the effective gradient abruptly shifts between
references during training. This leads to noisy, highly variable updates and makes
MDPO sensitive to small fluctuations in the reference log-probabilities, even when
log-likelihoods are normalized and computed with \texttt{logsumexp}.

\end{document}